\documentclass[nohyperref]{article}

\usepackage{microtype}
\usepackage{graphicx}
\usepackage{subcaption}
\usepackage{booktabs} 

\usepackage{hyperref}

\usepackage[accepted]{icml2022_pods}

\usepackage{amsmath}
\usepackage{amssymb}
\usepackage{mathtools}
\usepackage{amsthm}

\usepackage[capitalize,noabbrev]{cleveref}

\theoremstyle{plain}

\theoremstyle{definition}

\theoremstyle{remark}

\usepackage[textsize=tiny]{todonotes}

\icmltitlerunning{Towards Practicable Sequential Shift Detectors}

\begin{document}

\twocolumn[
\icmltitle{Towards Practicable Sequential Shift Detectors}




\begin{icmlauthorlist}
\icmlauthor{Oliver Cobb}{yyy}
\icmlauthor{Arnaud Van Looveren}{yyy}
\end{icmlauthorlist}

\icmlaffiliation{yyy}{Seldon Technologies}

\icmlcorrespondingauthor{Oliver Cobb}{oc@seldon.io}

\icmlkeywords{Machine Learning, ICML}

\vskip 0.3in
]



\printAffiliationsAndNotice{}  

\begin{abstract}
There is a growing awareness of the harmful effects of distribution shift on the performance of deployed machine learning models. Consequently, there is a growing interest in detecting these shifts before associated costs have time to accumulate. However, desiderata of crucial importance to the practicable deployment of sequential shift detectors are typically overlooked by existing works, precluding their widespread adoption. We identify three such desiderata, highlight existing works relevant to their satisfaction, and recommend impactful directions for future research.
\end{abstract}

\section{Introduction}\label{sec:intro}

Machine learning models are increasingly trusted to make decisions of real world consequence. If the distribution underlying deployment data matches that of the training data then model performance can be expected to match that observed on validation data. But if the distribution changes, shift is said to have occurred and model performance can suffer catastrophically \cite{taori2020measuring, ovadia2019can}. It is therefore important to have systems in place that test for such changes sequentially, as the instances arrive.

There is a rich literature in topics highly relevant to designing such systems, such as change detection \cite{basseville1993detection} and two-sample testing \cite{lehmann2005testing, salmaso2010permutation}. However, there is relatively little work that transfers ideas therein to the design of sequential shift detectors practicable for real world deployment scenarios. In such scenarios labels are typically unavailable, making it necessary to look for change in the feature space. These features often have a complex multivariate form (e.g. images or text) that requires projection into a lower dimensional space suitable for applying a two-sample test. However specifying such a projection prior to observing the data being tested is typically impossible.

Even if such a projection can be specified, strict tests of equality are unrealistic. Training data is typically much more varied than a small batch of recent deployment data. Whilst this deviation from the i.i.d.\ assumption is ostensibly what shift detectors are designed to detect, in practice the assumption is often broken in expected, permissible and unavoidable ways. In such cases, in order for monitoring systems to be at all useful it must be possible for practitioners to declare insensitivity to these expected changes.

Unlike in many traditional works \cite{gama2004learning, baena2006early} where detections automatically trigger cheap and simple adaptation/retraining procedures, modern machine learning models can be difficult and expensive to adapt. Practitioners therefore require the ability to specify precisely how often they are willing to incur the cost of false detections. Thankfully, unlike in traditional change detection settings, there exists a large set of data from the pre-change distribution: the training data. How best to use this data to configure detectors to operate with known behaviour in the absence of change is underexplored.


Consider, as a harmonising example, deploying a model to identify skin lesions. We wish to detect unexpected data shift, due to equipment malfunction for example, as quickly as possible in a sequential manner. The detector may access images and associated patient metadata, but not associated labels: indeed if they were available they would not need predicting. Each detection requires a manual response, necessitating an ability to specify an acceptable rate of false detections. We do not wish to operate at some unknown rate below this and increase the risk of actual shifts going undetected. We happen to know that the prevalences of different age groups vary significantly throughout the day. This is not a problem to the model, which has been trained on patients of all ages. However it means that batches of recent deployment data deviate from the i.i.d.\ assumption in an expected way. We wish to only detect shifts that can not be attributed to a change in the distribution of ages.

Despite its typicality, several of the desiderata critical to this problem are overlooked by the vast majority of existing works. Following a brief review of sequential shift detection in Section~\ref{sec:background}, in Section~\ref{sec:desiderata} we discuss these desiderata and highlight works relevant to their satisfaction. In Section~\ref{sec:future} we identify further impactful directions for future research.

\section{Background}\label{sec:background}

Let $M$ denote a model mapping features $x \in \mathcal{X}$ onto labels $y \in \mathcal{Y}$. Let $(X,Y) \in \mathcal{X}^n \times \mathcal{Y}^n$ denote reference data from the distribution 
$p(x,y)$ underlying the data on which $M$ was trained\footnote{If using the model within the shift detection procedure then this reference data should not be the split used for training.}. Let $((x_t,y_t))_{t\geq 1}$ denote the stream of data on which $M$ is deployed, with underlying distributions $(p_t)_{t\geq 1}$. The sequential shift detection problem can then be described as detecting the time step at which deployment samples stop constituting i.i.d.\ samples from $p$. Typically, the sudden change point formalisation of
\vspace{-0.6mm}
\begin{equation}\label{eqn:model}
    p_t =
        \begin{cases}
            p & \text{for $t < \tau$} \\
            q & \text{for $t \geq \tau$},
        \end{cases}
\end{equation}
is adopted, where $p$ and $q$ denote pre- and post-change distributions and $\tau \in \mathbb{N}$ an unknown change point. Letting $T$ denote the time at which a detection is made, algorithms are designed to achieve a small\footnote{Precise objectives relating to the random variable $T - \tau$ vary.} delay $T - \tau$, subject to constraints on the distribution of $T$ in the absence of change.
These competing objectives represent a classic trade off of statistical power against false positive rate control. For a given detection algorithm the distribution of $T$ in the absence of change is independent of any post-change distribution $q$ or change point $\tau$, which can be thought to be at infinity. Let $T_{\infty}$ denote the corresponding random variable.

Sometimes particular types of shift are of interest. The case $p(y|x)=q(y|x)$ but $p(x)\neq q(x)$ is referred to as \textit{covariate shift}, whilst $p(x|y)=q(x|y)$ but $p(y)\neq q(y)$ is referred to as \textit{label shift}. When the causal structure of the problem is such that shift can be assumed to take one of these forms \cite{scholkopf2012causal}, the model can be updated using only unlabelled data from the post-change distribution \cite{huang2006correcting, sugiyama2007direct, saerens2002adjusting, lipton2018detecting}. More generally, however, one should assume the shift affects the relationship between $x$ and $y$. In such cases \textit{concept shift} is said to have occurred. Often one is only interested in detecting \textit{malicious shift}: that which causes an undesirable change to the distribution of $\ell(y, M(x))$, where $\ell$ is a model performance metric of interest. Covariate, label and concept shift can all be malicious.

A variety of sequential shift detectors have been proposed, some specifically targeting one of the subcategories described above. The vast majority, however, can be described within the framework described by \citet{lu2018learning}. Within this framework, at each time step four stages are performed:
\vspace{-5mm}
\begin{enumerate}
    \item A window of recent deployment data $(\tilde{X}, \tilde{Y})$ is selected for comparison to the reference data $(X,Y)$.
    \item A summary statistic $s: \mathcal{X} \times \mathcal{Y} \rightarrow \mathcal{S}$ is used to project the data onto summaries $S$ and $\tilde{S}$.
    \item The value of a test statistic $\hat{d}(S, \tilde{S})$ is computed.
    \item A detection decision is made, typically by comparing the test statistic $\hat{d}(S, \tilde{S})$ to a threshold $\hat{h}$.
\end{enumerate}
\vspace{-3mm}
Although the four stages can be described and reasoned about separately, in practice the boundaries are fuzzy and the most effective algorithms are designed such that the stages operate effectively together, both computationally and statistically. In particular, it is usually necessary for the test statistic to be updated in a sequential manner as the deployment window is updated with new data.

Targeting specific types of shift can be achieved through specification of the summary statistic $s$ and test statistic $\hat{d}$. For example, denoting $S=\{s_i\}_{i=1}^n$ and $\tilde{S}=\{\tilde{s}_j\}_{j=1}^m$, if one is interested in detecting a decrease in the average model loss then one would choose $s(x,y)=\ell(y,M(x))$ and $\hat{d}(S, \tilde{S}) = \frac1n \sum_{i=1}^n s_i - \frac1m \sum_{j=1}^m \tilde{s}_j$. This captures the essence of early approaches to shift detection \cite{gama2004learning, baena2006early, bifet2007learning}. However, a major restriction that prevents the application of these and many similar approaches is that deployment labels $\tilde{Y}$ are not usually available at detection time, necessitating summary statistics that depend only on $x$. This means that shift that is purely in the conditional distribution $p(y|x) \neq q(y|x)$ and not the covariate distribution $p(x)=q(x)$ is unidentifiable. Nevertheless, unless there is reason to believe this invariance holds, the weaker assumption is that the difference between $p$ and $q$ is identifiable through that between $p(x)$ and $q(x)$. Furthermore, the absence of deployment labels does not restrict to detecting only covariate shift, particularly when assumptions can be made. For example if the $p(x|y)=q(x|y)$ assumption can be made then $s(x,y)=M(x)$ makes a suitable summary statistic for detecting label shift \cite{lipton2018detecting}. Alternatively, under a $p(y|x)=q(y|x)$ assumption one can target covariate shift likely to be malicious by using the labelled training data to identify regions of the covariate space on which the model struggles to perform and constructing a test statistic to penalise shift into these regions \cite{sethi2017reliable, baier2021detecting}. However, absent the ability to make such assumptions, one must typically resort to the most general available summary, $s(x,y)=x$.

Whilst there are various test statistics suitable for multivariate data, such as those based on estimating the maximum mean discrepancy (MMD) \cite{gretton2012mmd} and least-squares density difference (LSDD) \cite{bu2016pdf} between underlying distributions, their effectiveness typically depends on the extent to which implicitly or explicitly defined distances capture a relevant notion of similarity. This makes the $s(x,y)=x$ case statistically challenging when $\mathcal{X}$ lacks a relevant associated metric. For example the Euclidean distance between images is not relevant if one is interested in detecting drift at a semantic level. It is therefore usually necessary to project the data into a latent space where distances are meaningful. The implicit assumption is that changes of interest remain identifiable. Hence there is a trade-off between choosing high dimensional representations most likely to preserve information relevant to the change and low dimensional representations in which distances are more meaningful and two-sample tests therefore more powerful. Whilst task- or model-specific projections are popular, they can be problematic as it should be assumed that they discard all information not discriminative under $p$, which could become discriminative under $q$.
 An alternative, described in more detail in Section~\ref{sec:desiderata}, is to outsource the problem of defining a space in which to look for differences by using a split of the available data to directly optimise a projection to separate reference from deployment samples.


\section{Overlooked Desiderata}\label{sec:desiderata}

The most obvious desideratum for sequential shift detection is that of statistical power: the detector should be much more likely to make a detection following a change than before. The vast majority of works prioritise this property and demonstrate empirically the extent to which they satisfy it. Another obvious desideratum is a low cost of processing each arriving instance. We now identify and discuss three additional desiderata that we believe do not receive as much research attention as their importance in practice warrants. We believe practitioners desire:
\vspace{-2mm}
\begin{itemize}
    \item[D1:] To configure detectors to operate with known behaviour in the absence of change.
    \item[D2:] To outsource the burden of identifying a discriminative test statistic.
    \item[D3:] Flexibility over which changes should -- and should not -- be detected as shift.
\end{itemize}

\subsection{Detector Calibration}




Unlike two-sample testing in the offline setting where only a single test is performed, for sequential shift detection an identical test is repeatedly performed using mostly the same data. Assuming a sliding deployment window containing the $w$ most recent observations, $n+w-1$ of the $n+w$ observations used for the test at time $t$ are the same as those used at time $t-1$. Consecutive test statistics are therefore highly correlated. In general there is no straightforward way to relate p-values associated with offline statistical tests with desired constraints on the distribution of the detection time $T$ in the absence of change.

Some works on sequential shift detection neglect to address the problem completely by treating detection thresholds as hyperparameters that practitioners can choose manually \cite{antwi2012perfsim, gozuaccik2019unsupervised}. In practice however, this renders the entire detection system unusable. A less obvious but more common way in which the problem is somewhat trivialised is by using a conventional (offline) estimate $\hat{h}$ of the threshold $h$ which the first test statistic exceeds with a prescribed false-positive probability $\alpha$ \cite{dos2016fast, bu2017incremental}. By comparing subsequent test statistics to the same threshold one can upper-bound the probability of a false detection at each time step. This is because knowledge that the first test statistic is below the threshold makes it more likely that the highly correlated test statistic that follows also is. The problem is that the tightness of such bounds is rarely considered, neither analytically or empirically, thereby providing practitioners with little knowledge of the distribution of $T_{\infty}$. We include some simple experiments in Appendix~\ref{sec:appendix} that show that resulting detectors operate at false positive rates orders of magnitude below that corresponding to the bound. Not only does this mean that the significance of detections are unknown when they occur, but also that the detector is operating with statistical power much below that which would be possible if operating under the practitioner's desired false positive rate. When such bounds are the only means of control there is no way for practitioners to trade off the costs of false detections with that of missed changes.

There are many works in the field of change detection that offer more control over the distribution of $T_{\infty}$, but the unavailability of a reference set in the typical change detection setting constrains the problem and makes additional assumptions necessary. For example, for univariate streams \citet{kifer2004detecting} leverage rank-based statistics whose distribution are independent of the pre- and post-change distributions to construct change detection algorithms that can be configured to operate such that $P(T_{\infty} \leq \lambda)$ is tightly upper-bounded by $\alpha$. \citet{ross2012two} similarly leverage rank-based statistics to instead configure detectors that operate such that $E[T_{\infty}]$ can be specified with high accuracy. \citet{vovk2021retrain} consider various ways of controlling $T_{\infty}$, with and without using a reference set from $p$, using exchangeability martingales that similarly require $s(x,y)$ to be univariate. Unfortunately, however, there is no straightforward generalisation of these rank- or exchangeability martingale-based statistics to the multivariate setting. 

Moreover, even if $P(T_{\infty} \leq \lambda)$ or $E[T_{\infty}]$ can be accurately specified, using a single threshold $\hat{h}$ across all time steps results in false-positive probabilities that vary significantly across time steps, complicating the interpretation of the significance of detections. \citet{verdier2008adaptive} provide an approach for setting time-varying thresholds $(\hat{h}_t)_{t \geq 1}$ such that $T_{\infty}$ approximately follows the memoryless distribution  $\text{Geom}(\alpha)$ for a desired $\alpha$. Their approach is not limited to univariate streams, but they instead require complete knowledge of the pre-change distribution $p$. Few works address this problem in the nonparametric multivariate case relevant to machine learning model monitoring. \citet{li2019scan} provide, for an MMD-based test statistic, an analytic estimator $\hat{h}$ of the fixed threshold $h$ achieving a desired $E[T_{\infty}]$ that is accurate in the asymptotic limit $h \rightarrow \infty$. \citet{cobb2022sequential} show that this estimator results in significant miscalibration in the finite regime and propose an alternative, test-statistic agnostic approach to setting time-varying thresholds in a simulation-based manner akin to \citet{verdier2008adaptive}. Their method is particularly well suited to test statistics for which the cost of computation over a large number of permutations of the reference data can benefit from significant amortisation, such as MMD- or LSDD-based statistics.

\subsection{Learning Discriminative Test Statistics}

A challenge inherent in the shift detection problem, even in the offline case, is that one does not know the way in which the pre- and post-change distributions $p$ and $q$ will differ. This leads one to consider test statistics $\hat{d}$ that estimate a metric or divergence $d$ between probability distributions, satisfying $d(p,q)=0 \iff p=q$. One such example is $d(p,q)=\text{MMD}_k(p,q)$, which is a metric assuming the kernel $k$ is universal, meaning any consistent estimator $\hat{d}(p,q)$ will correctly identify a change given enough samples. In practice however only limited samples may be used for estimation and the power of the statistic depends on the extent to which the kernel's reproducing kernel Hilbert space (RKHS) $\mathcal{H}_k$ is suited to identifying differences between $p$ and $q$. For complex data often of interest in machine learning contexts manually specifying a suitable kernel is not plausible. \citet{liu2020learning} demonstrate the effectiveness of using a portion of the samples to learn such a kernel and then using the remaining samples to perform the statistical test. \citet{liu2020learning} note that this approach closely parallels those that simply train a binary domain classifier to differentiate reference from deployment samples \cite{lopezpaz2017revisiting}. If instances unseen by the domain classifier can be classified as reference or deployment instances, with accuracy significantly better than chance, then a difference between the underlying distributions must exist. 
There has been interest in learning discriminative test statistics in a manner that does not require separate splits for learning and testing \cite{kubler2020learning, schrab2021mmd}, however these approaches are currently limited to selecting combinations of test statistics from a predefined set.

Despite the power of two-sample tests based on learned test statistics we are not aware of any work that aims to transfer their success to the sequential detection setting. 
Such an approach would likely partition the stream into a substream for updating an online binary classifier (or alternative discriminative representation) and a substream on which to evaluate how well it generalises.  Whilst we forsee possible challenges around the robustness of the updating scheme and achieving calibration within the overarching detector, we believe this area to be fertile ground for future research.

\subsection{Permitting Certain Types Of Change}

For many real world shift detection problems even well calibrated detectors that powerfully detect changes are of little use. Often a window of the most recent deployment data is not expected to form an i.i.d.\ sample from the distribution underlying the reference data. Consider the example, from \citet{cobb2022context}, of a vision model trained to classify images of animals. The distribution underlying the images changes throughout the day. For example the distribution underlying a nighttime batch of deployment images differs from that underlying the reference set, which additionally contains daytime images. A test of equality would therefore make a detection at every time step, making it impossible to detect unexpected changes of actual interest. Note that even if labels are available, this problem is not solved by monitoring an indicator of model performance $\ell(y,M(x))$: the model performance might be expected to vary in a predictable and expected way depending on the lighting conditions. We believe there is a lack of work exploring how to afford practitioners the ability to specify which changes should -- and should not -- be detected.

\citet{cobb2022context} propose one possible framework for allowing this flexibility. Alongside the usual summary statistic $s(x,y)$ their framework allows the specification a context variable $c$, which may be a deterministic transformation of $(x,y)$ or related in some unknown probabilistic manner. Then only differences between summaries $S$ and $\tilde{S}$ that cannot be attributed to differences between corresponding contexts $C$ and $\tilde{C}$ cause a detection to be made. They propose a kernel-based instantiation of their framework based on the approach of \citet{park2021conditional} for detecting conditional distributional treatment effects.

\section{Future Work}\label{sec:future}

There is a tendency in sequential shift detection research to focus on just a portion of the problem, perhaps corresponding to one or two of the four stages outlined in Section~\ref{sec:background}, whilst trivialising the importance of the others. For example approaches based on repeatedly applying established multivariate tests trivialise the projection onto a space suitable for such tests. Similarly stage 4 is often trivialised by methods that only allow false positive control through extremely slack bounds. Before detectors can become widely deployed alongside machine learning models there is a need for methods addressing the problem in a holistic end-to-end manner that satisfies a wider set of desiderata. This will require the development of new frameworks that will likely require components such as summary and test statistics be constrained to satisfy certain properties. The establishment of such frameworks should take priority before further research can then identify instantiations empirically effective with respect to established desiderata such as statistical power.

\bibliography{main}
\bibliographystyle{icml2022}

\newpage
\appendix
\onecolumn
\section{Appendix}\label{sec:appendix}
\subsection{Slackness Resulting from Correlated Test Statistics}

In this section we briefly highlight the problem that results from a failure to account for the correlation between consecutive test statistics. First recall the four stage framework discussed in Section~\ref{sec:background}. In particular recall that whether a shift detector makes a detection at time step $t$ typically depends on whether a corresponding test statistic $\hat{d}(S,\tilde{S}_t)$ exceeds a threshold $\hat{h}_t$, where $\tilde{S}_t$ denotes the window of deployment summaries at time step $t$. Most commonly, a sliding window of a fixed size $w$ is used such that $\tilde{S}_t = \{s(x_j,y_j)\}_{j=t-w+1}^{j=t}$. 

An approach that is particularly prevalent in the literature is to use a time-invariant threshold ($\hat{h}_t=\hat{h}$ for $t=1,2,...$) that estimates the threshold $h$ exceeded by the first test statistic $\hat{d}(S,\tilde{S}_W)$ with a prescribed false positive probability $\alpha$. Note that this is equivalent to estimating p-values at each time step and making a detection if one falls below $\alpha$. If consecutive test statistics were statistically independent, and $h$ accurately estimated, this results in a detector for which the expected run time to false detection $E[T_\infty]$ approximately equals $1/\alpha$. However the correlation between test statistics means that instead it results in detectors for which $1/\alpha$ is a lower bound on $E[T_{\infty}]$. Whilst this lower bounding can indeed be considered as a form of ``control" over the false positive rate, we have found from practical experience that it frequently results in the configuration of detectors that operate at such a low false detection rate that the true detection rate is also unknowingly unacceptably low. We provide some very simple experiments that demonstrates how slack the bound can become in practice.

\begin{figure}[!b]
    \centering
    \begin{subfigure}[b]{0.4\textwidth}
        \centering
        \includegraphics[width=\textwidth]{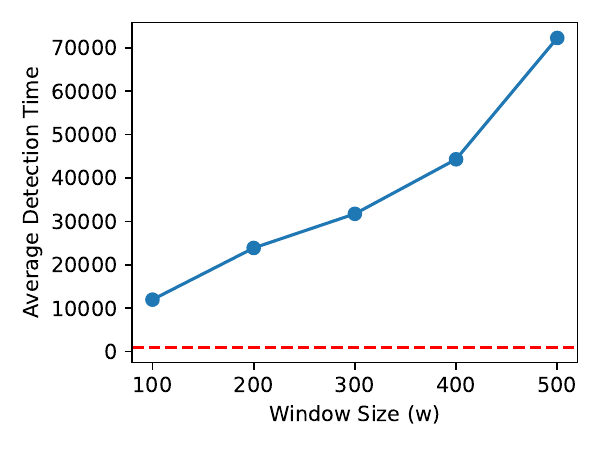}
        \caption{Average detection times for a range of window sizes $w$ and reference set size fixed at $n=3000$.}
        \label{fig:a}
    \end{subfigure}
    \hspace{2cm}
    \begin{subfigure}[b]{0.4\textwidth}
        \centering
        \includegraphics[width=\textwidth]{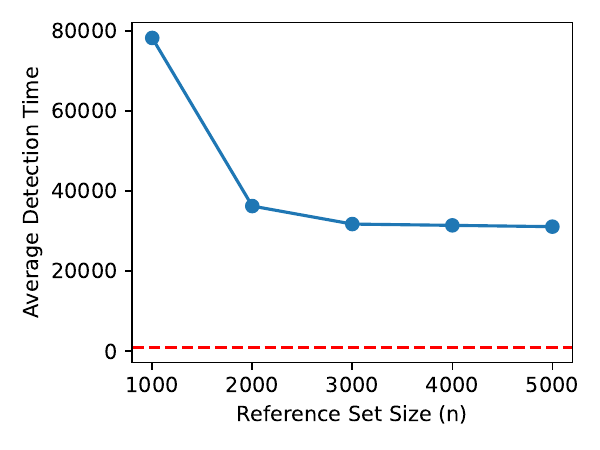}
        \caption{Average detection times for a range of reference set sizes $n$ and a window size fixed at $w=300$.}
        \label{fig:b}
    \end{subfigure}
       \caption{Average run time to false detection for a range of reference set sizes $n$ and window sizes $w$. The desired expected run time to false detection of 1000 is shown as a red dotted line.}
       \label{fig:both}
\end{figure}

We consider the simple setting where the distribution of the summary statistics is simply the standard normal distribution $N(0,1)$, for both the reference set $S$ of size $n$ and deployment stream $(s_t)_{t \geq 1}$. We consider a deployment window $\tilde{S}_t = \{s(x_j,y_j)\}_{j=t-w+1}^{j=t}$ updated at each time step $t$ to contain the $w$ most recent statistics. As the test statistic we (similarly to \citet{dos2016fast}) use the Kolmogorov-Smirnov distance $\hat{d}(S,\tilde{S}_t)=\max_u|F_n(u)-G_{w,t}(u)|$, where $F_n$ and $G_{w,t}$ are the empirical cumulative distribution functions of $S$ and $\tilde{S}_t$ respectively. This allows, for a prescribed false positive probability $\alpha$, the accurate estimation of $h$ using standard methods \cite{hodges1958significance}. We wish to specify an expected run time to false detection $E[T_\infty]$  of 1000, and therefore specify $\alpha=1/1000$. Figure \ref{fig:both} shows, over 250 independent runs, the actual average run time to false detection for a number of reference set sizes and window sizes.

We see from Figure~\ref{fig:a} that for a moderately sized deployment window of $w=100$ and reference set size of $n=3000$, the actual $E[T_{\infty}]$ is already 11 times larger than the lower bound. As the window size increases, and therefore the correlation between consecutive statistics grows, the slackness increases. At a window size of $w=500$ the actual $E[T_{\infty}]$ is 72 times larger than the lower bound. Increasing the window size further would of course result in further slackness, but is beyond the computational budget of these experiments. By contrast, increasing the reference set size does not affect the correlation between consecutive test statistics, but does affect the ability to obtain statistically significant ($\alpha=0.001$) values. Figure~\ref{fig:b} shows that the actual $E[T_{\infty}]$ therefore initially decreases with increasing reference set size but then levels off to approximately 32 times the lower bound (for a window size of $w=300$).

The takeaway from these experiments is that although the described approach can be used to specify a lower bound on the expected run time to false detection $E[T_{\infty}]$ in the absence of change, in practice it will actually take some unknown value likely orders of magnitude above this lower bound that depends on various factors such as reference set size, window size, test statistic and pre-change distribution. This makes it impossible for the approach to be used to correspond to constraints on how often practitioners are willing to respond to false detections. Thresholds need to be set using approaches that account for the correlation between consecutive test statistics. Other than the recent work of \citet{cobb2022sequential}, there is relatively little work exploring how this can be achieved in the general (multivariate and nonparametric) case, despite the availability of the large set of reference data $S$ from the pre-change distribution.

\end{document}